\documentclass{llncs}
\usepackage{llncsdoc}
\usepackage{graphicx}
\usepackage{cite}
\usepackage{hyperref}
\usepackage[T1]{fontenc}
\usepackage[utf8]{inputenc}     
\usepackage[english]{babel}   
\usepackage{float}

\newcommand{\repeatthanks}{\textsuperscript{\thefootnote}}

\begin{document}
\title{Dataset Augmentation with Synthetic Images Improves Semantic Segmentation}
\author{Manik Goyal\inst{1}\thanks{Authors contributed equally to this manuscript}, Param Rajpura \inst{2}\repeatthanks, Hristo Bojinov \inst{3} \and Ravi Hegde \inst{2}}

\institute{Indian Institute of Technology (BHU) Varanasi, India 221005
\and Indian Institute of Technology Gandhinagar, India 382355
\and Innit Inc., USA 94063}
\maketitle
%
%\newpage
%\tableofcontents
%\newpage
%
\begin{abstract}
Although Deep  Convolutional Neural Networks trained with strong pixel-level annotations have significantly pushed the performance in semantic segmentation, annotation efforts required for the creation of training data remains a roadblock for further improvements.  We show that augmentation of the weakly annotated training dataset with synthetic images minimizes both the annotation efforts and also the cost of capturing images with sufficient variety. Evaluation on the PASCAL 2012 validation dataset shows an increase in mean IOU from 52.80\% to 55.47\% by adding just 100 synthetic images per object class. Our approach is thus a promising solution to the problems of annotation and dataset collection. 
\end{abstract}
\section{Introduction}
Deep Convolutional Neural Networks (CNNs) have achieved state-of-the-art performance on several image processing and computer vision tasks like image
classification, object detection, and segmentation~\cite{Krizhevsky2012,Szegedy2015}. Numerous applications depend on the ability to infer knowledge about the environment through image acquisition and processing. Hence, scene understanding as a core computer vision problem has received a lot of attention. Semantic segmentation, the task of labelling pixels by their semantics (like 'person', 'dog', 'horse'), paves the road for complete scene understanding. Current state-of-the-art methods for semantic segmentation are dominated by deep convolutional neural networks (DCNNs)~\cite{Shelhamer2016,Badrinarayanan2015,Chen2016}. However, training end-to-end CNNs requires large scale annotated datasets. Even with a large enough dataset, training segmentation models with only image level annotations is quite challenging~\cite{Vezhnevets2012,Verbeek2007,xu_cvpr2014} as the architecture needs to learn from higher level image labels and then predict low-level pixel labels. The significant problem here is the need for pixel-wise annotated labels for training, which becomes a time-consuming and expensive annotation effort.The Pascal Visual Object Classes (VOC)~\cite{Everingham2015} challenge considered to be a standard dataset for challenges like classification,
detection, segmentation, action classification, and person
layout provides only 1464 (training) and 1449 (validation) pixel-wise labelled images for semantic segmentation challenge. Some researchers have extended training dataset~\cite{Hariharan2011} with 8.5k strong pixel-wise annotations (consisting of the same 21 classes as PASCAL VOC) to counter this problem. In practical applications, the challenge still stands since various classes of objects need to be detected, and annotated dataset for such training is always required. 

To reduce the annotation efforts, recent reports use weakly-annotated datasets to train deep CNN models for semantic segmentation. Typically, such weak annotations take the form of bound-boxes, because forming bound-boxes around every instance of a class is around 15 times faster than doing pixel-level labelling~\cite{Lin}. 
These approaches rely on either defining some constraints~\cite{Pathak2015} or on using multiple instance learning~\cite{Pathak2014} techniques.~\cite{Rother2004} uses GraphCut for approximating bound-boxes to semantic labels. Although deep CNNs (such as the one proposed by~\cite{Papandreou2015} using DeepLab model~\cite{Chen2016}) significantly improved the segmentation performance using such weakly-annotated datasets, they failed to provide good visualization on test images.

There have been solutions proposed to reduce annotation efforts by employing transfer learning or simulating scenes. The research community has proposed multiple approaches for the problem of adapting vision-based models trained in one domain to a different domain~\cite{Li2014,Hoffman2013,Hoffman2014,Kulis2011,Long2015}. Examples include:  re-training a model in the target domain~\cite{Yosinski2014}; adapting the weights of a pre-trained model~\cite{li2017revisiting}; using pre-trained weights for feature extraction~\cite{Gupta2016}; and, learning common features between domains~\cite{Tzeng2014}. Augmentation of datasets with synthetically rendered images or using datasets composed entirely of synthetic images is one of the techniques that is being explored to address the dearth of annotated data for training all kinds of CNNs. Significant research in transfer learning from synthetically rendered images to real images has been published~\cite{RosCVPR16,Peng2017}. Most researchers have used gaming or physics rendering engine to produce synthetic images~\cite{RosCVPR16} especially in the automotive domain.  Peng et al.~\cite{Peng2015} have done progressive work in the Object Detection context, understanding various cues affecting transfer learning from synthetic images. But they train individual classifiers for each class after extracting features from pre-trained CNN. They show that adding different cues like background, object textures, shape to the synthetic image increases the performance~\cite{Peng2016,Peng2017} for object detection. There has not been an attempt yet to benchmark performances on the standard PASCAL VOC~\cite{Everingham2015} semantic segmentation benchmark using synthetic images.

To the best of our knowledge, our report is the first attempt at combining weak annotations (generating semantic labels from bound box labels) and synthetically rendered images from freely available 3D models for semantic segmentation. We demonstrate a significant increase in segmentation performance (as measured by the  mean of pixel-wise intersection-over-union (IoU))  by using semantic labels from weak annotations and synthetic images. We used a Fully Convolutional Network (FCN-8s) architecture~\cite{Shelhamer2016} and evaluate it on the standard PASCAL VOC~\cite{Everingham2015} semantic segmentation dataset. The rest of this paper is organized as follows: our methodology is described in
~\autoref{sec:method}, followed by the results we obtain reported in
~\autoref{sec:results}, finally concluding the paper in~\autoref{sec:conclusion}.
\section{Method} \label{sec:method}

\begin{figure}[htbp]
\begin{center}
\includegraphics{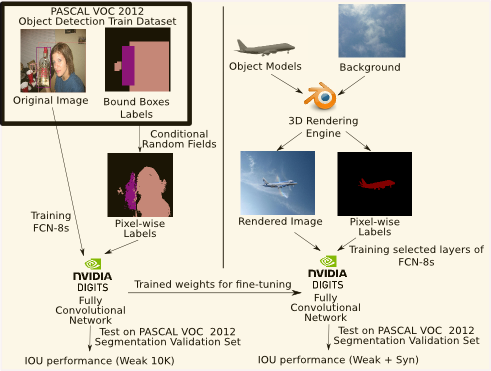}
\end{center}
   \caption{Overview representing the approach for learning semantic segmentation from weak bound-box labels and synthetic images rendered using 3D models.}
\label{fig:Overview}
\end{figure}

Given an RGB image capturing one or many of the 20 objects included in PASCAL VOC 2012 semantic segmentation challenge, our goal is to predict a
label image with pixel-wise segmentation for each object of interest. Our approach, represented in~\autoref{fig:Overview}, is to train a deep CNN with synthetic rendered images from available 3D models.  We divide the training of FCN into two stages: fine-tuning FCN with the \texttt{Weak(10k)} dataset (real images with bound box labels) generated from bound box labelled images, and fine-tuning with our own \texttt{Syn(2k)} dataset (synthetic images rendered from 3D models). Our methodology
can be divided into two major parts: dataset generation and fine-tuning of the FCN.  These are explained in the following subsections. 

\subsection{Dataset generation}

\begin{figure}[htbp]
\begin{center}
{ \includegraphics[width=0.9\linewidth]
{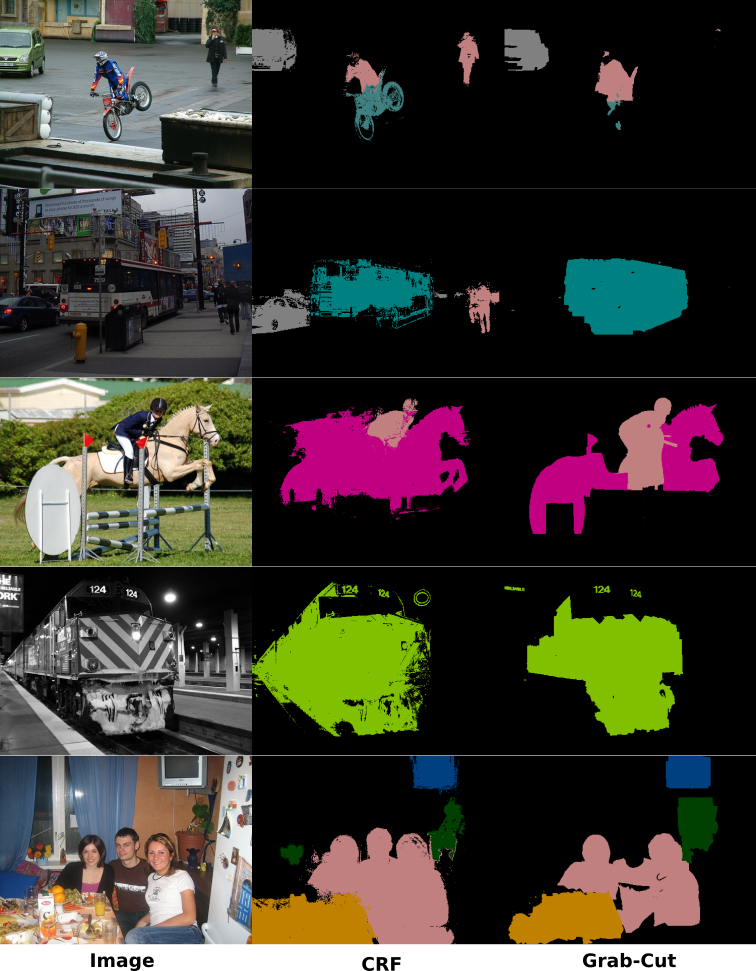}}
\end{center}
   \caption{Comparision of CRF and Grab-Cut segmentation from bound box labels}
\label{fig:crf}
\end{figure}

\textbf{Weakly supervised semantic annotations:} To train the CNN for semantic segmentation, we use the available bound-box annotations in PASCAL-VOC object detection challenge training set (10k images with 20 classes). Since the bound-boxes fully surround the object including pixels from background, we filter those pixels into foreground and background. Later the foreground pixels are given their corresponding object label in cases where multiple objects are present in an image.
Two methods were chosen for converting bound-boxes to semantic segmentation  namely Grab-Cut~\cite{Rother2004} and Conditional Random Fields (CRF) as deployed by ~\cite{Chen2016,Papandreou2015}. Based upon the performance on a few selected images, we use the labels from CRF for training the CNN. ~\autoref{fig:crf} shows the comparison of results from both methods. Grab-Cut tends to miss labelling smaller objects but is precise in labelling larger objects. CRF labels objects of interest accurately with a small amount of noise around the edges.

\textbf{Synthetic images rendered from 3D models:} We use the open source 3D graphics software Blender for this purpose. Blender-Python APIs facilitates the  loading of 3D models and automation of scene rendering.
We use Cycles Render Engine available with Blender since it supports ray-tracing to render synthetic images. 
Since all the required information for annotation is available, we
use the PASCAL Segmentation label format with labelled pixels for 20 classes.

Real world images have lot of information embedded about the environment, illumination, surface materials, shapes etc. Since the trained model, at test time must be able to generalize to the real world images, we take into
consideration the following aspects during generation of each scenario:
\begin{enumerate}
  \item Number of objects
  \item Shape, Texture, and Materials of the objects
  \item Background of the object
  \item Position, Orientation of camera
  \item Illumination via light sources
\end{enumerate}

\begin{figure}[htbp]
\begin{center}
{ \includegraphics[width=\linewidth]
{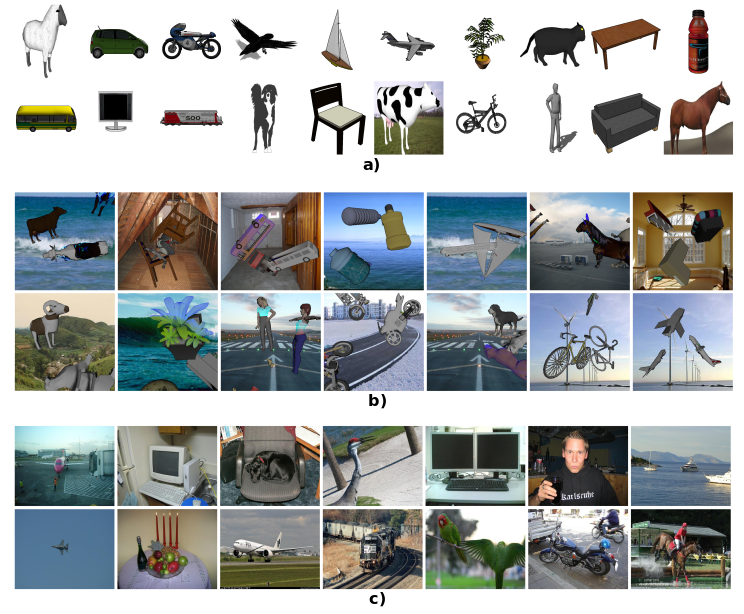}}
\end{center}
   \caption{(a) 3D models used for synthetic dataset (b) Synthetic images with multiple objects (c) Images from the training set of PASCAL VOC 2012 Object Detection Dataset}
\label{fig:TrainSamples}
\end{figure}

In order to simulate the scenario, we need 3D models, their texture information
and metadata. Thousands of 3D CAD models are available online. We choose
ShapeNet~\cite{Chang2015} database since it provides a large variety of models in the 20 categories for PASCAL segmentation challenge. ~\autoref{fig:TrainSamples}a
shows few of the models used for rendering images. The variety helps randomize the aspect of shape, texture and materials of the objects. We use images from SUN database~\cite{Xiao2010} as background images. From the large categories of images, we select few categories relevant as background to the classes of objects to be recognized.

For generating training set with rendered images, the 3D scenes need to be
distinct. For every class of object, multiple models are randomly chosen from the model repository. Each object is scaled, rotated and placed at random location within the field of view of the camera which is placed at a pre-defined location. The scene is illuminated via directional light source.
Later, a background image is chosen from the database and the image is rendered with Cycles Render Engine, finally generating RGB image and pixel-wise labelled image.
~\autoref{fig:TrainSamples}b shows few rendered images used as training set while ~\autoref{fig:TrainSamples}c shows the subset of real images from PASCAL Object Detection dataset (\texttt{Weak(10k)}) used in training.
\begin{figure}[htbp]
\begin{center}
{ \includegraphics[width=0.9\linewidth]
{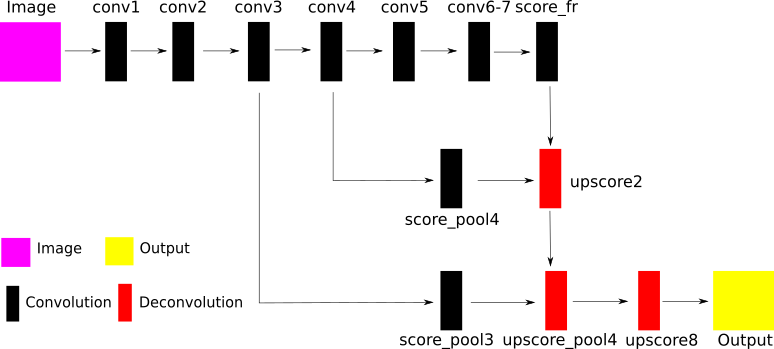}}
\end{center}
   \caption{FCN-8s Architecture~\cite{Shelhamer2016}}
\label{fig:FCN}
\end{figure}

\subsection{Fine-tuning the Deep CNN}
We fine-tune FCN-8s~\cite{Shelhamer2016} pretrained on ImageNet~\cite{JiaDeng2009} initially with 10k real images along with semantic labels generated from bound-boxes using CRF. All layers in the network are fined-tuned with base learning rate of ${1e^{-5}}$. We further reference this model as baseline model.
In next stage, we fine tune the baseline model with synthetic images generated from Blender. Selected layers (score\_pool3, score\_pool4, upscore2, upscore\_pool4 and upscore8 shown in ~\autoref{fig:FCN})  consisting of  2 convolutional and 3 deconvolutional layers are fine-tuned with base learning rate of ${1e^{-6}}$. The network is trained with Adam optimizer for pixel-wise softmax loss function. Since the rendered images from 3D models are not rich in terms cues like textures, shadows and hence are not photo-realistic, we choose to fine-tune only few layers to capture majorly the higher hierarchical features like shape of the object.

\section{Results and Discussion}\label{sec:results}

\begin{table*}[t]
\caption{IoU Values for different classes evaluated on the PASCAL Semantic Segmentation Validation Set~\cite{Everingham2015}. IoU Values higher than baseline model have been highlighted}
\resizebox{\textwidth}{!}{%
\begin{tabular}{|l|c|c|c|c|c|c|}
\hline
 & \textbf{Real(1.5k)} & \textbf{\begin{tabular}[c]{@{}c@{}}Weak(10k)\\ (Baseline)\end{tabular}} & \textbf{\begin{tabular}[c]{@{}c@{}}Weak(10k)+\\ Syn\_Car(100)\end{tabular}} & \textbf{\begin{tabular}[c]{@{}c@{}}Weak(10k)+\\ Syn\_Bot(100)\end{tabular}} & \textbf{\begin{tabular}[c]{@{}c@{}}Weak(10k)+\\ Syn\_Aero(100)\end{tabular}} & \textbf{\begin{tabular}[c]{@{}c@{}}Weak(10k)+\\ Syn(2k)\end{tabular}} \\ \hline
Background & \textbf{87.88} & 87.12 & \textbf{87.38} & 86.78 & 85.39 & \textbf{87.88} \\ \hline
\textbf{Aeroplane} & 58.07 & 68.81 & \textbf{69.75} & 67.98 & 67.89 & \textbf{69.39} \\ \hline
\textbf{Bicycle} & \textbf{46.47} & 27.39 & \textbf{27.48} & 24.29 & 23.13 & \textbf{27.65} \\ \hline
\textbf{Bird} & 54.61 & 55.46 & 55.35 & \textbf{57.65} & \textbf{63.90} & \textbf{57.23} \\ \hline
\textbf{Boat} & 39.73 & 42.76 & \textbf{46.25} & \textbf{45.62} & \textbf{48.89} & \textbf{47.76} \\ \hline
\textbf{Bottle} & 41.56 & 47.25 & 44.38 & \textbf{57.55} & 27.83 & \textbf{55.52} \\ \hline
\textbf{Bus} & 61.67 & 73.96 & \textbf{77.70} & 73.15 & 68.51 & \textbf{76.19} \\ \hline
\textbf{Car} & 48.47 & 51.15 & \textbf{69.31} & 50.42 & 47.04 & \textbf{61.83} \\ \hline
\textbf{Cat} & 64.93 & 71.37 & 69.32 & 68.80 & 65.39 & \textbf{72.58} \\ \hline
\textbf{Chair} & \textbf{16.92} & 13.17 & 11.35 & 10.38 & 5.29 & \textbf{14.91} \\ \hline
\textbf{Cow} & 30.43 & 57.88 & 57.28 & \textbf{58.98} & 56.24 & \textbf{59.23} \\ \hline
\textbf{Dining Table} & 39.47 & 42.95 & 42.93 & \textbf{42.98} & 27.21 & \textbf{48.50} \\ \hline
\textbf{Dog} & 53.43 & \textbf{60.74} & 59.65 & 58.71 & 54.36 & 60.56 \\ \hline
\textbf{Horse} & 44.40 & 54.92 & 54.34 & 54.70 & 51.41 & \textbf{56.64} \\ \hline
\textbf{Motor Bike} & 59.03 & 54.99 & \textbf{56.47} & 50.58 & 44.46 & \textbf{62.49} \\ \hline
\textbf{Person} & 65.47 & \textbf{66.01} & 65.70 & 63.92 & 61.50 & 63.43 \\ \hline
\textbf{Potted Plant} & 14.06 & 41.38 & 36.12 & \textbf{41.46} & 23.72 & \textbf{49.54} \\ \hline
\textbf{Sheep} & 56.05 & 65.71 & \textbf{66.08} & \textbf{66.27} & 61.90 & \textbf{65.87} \\ \hline
\textbf{Sofa} & 24.75 & 30.37 & 29.46 & 28.46 & 17.60 & \textbf{36.95} \\ \hline
\textbf{Train} & 54.05 & 67.48 & 66.47 & 64.04 & 56.34 & \textbf{68.69} \\ \hline
\textbf{Tv Monitor} & \textbf{38.62} & \textbf{27.99} & \textbf{28.26} & 27.97 & 16.73 & 22.06 \\ \hline
\textbf{Mean IoU} & 47.62 & 52.80 & \textbf{53.38} & 52.41 & 46.42 & \textbf{55.47} \\ \hline
\end{tabular}
}
\label{table}
\end{table*}

\begin{figure}[htbp]
\begin{center}
{\includegraphics[width=0.85\linewidth]{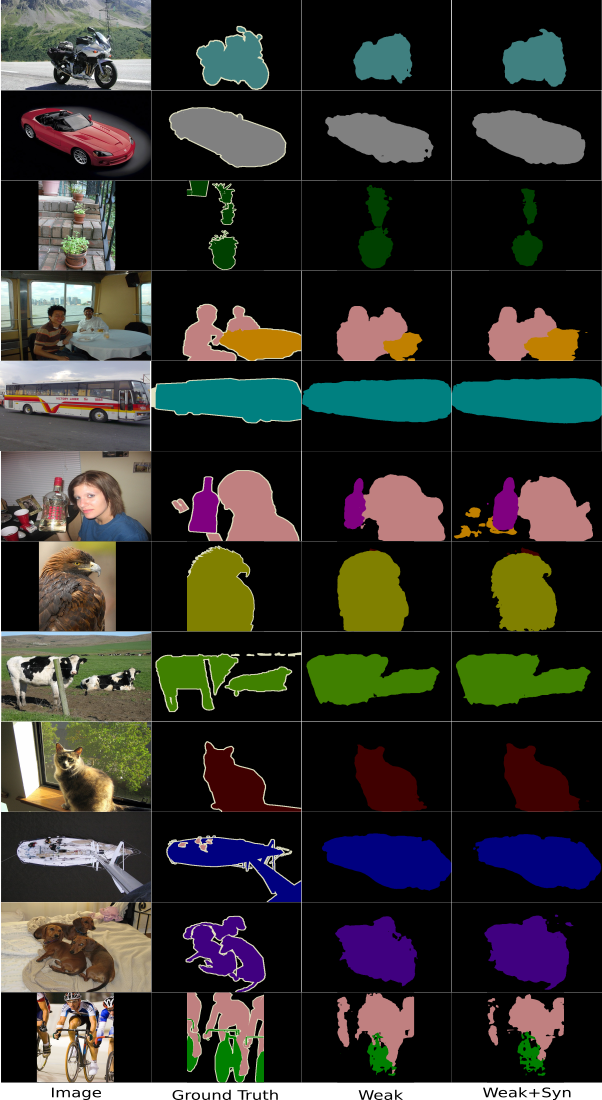}}
\end{center}
   \caption{Comparision of the segmented labels from Ground Truth, predictions from CNN trained with \texttt{Weak(10k)} and \texttt{Weak(10k)+Syn(2k)}}
\label{fig:Comp}
\end{figure}

The experiments were carried out using the workstation with Intel Core i7-5960X processor accelerated by NVIDIA GEFORCE GTX 1070. NVIDIA-DIGITS$^{TM}$ (v5.0)~\cite{NVIDIA} was used with Caffe library to train and manage the datasets.
The proposed CNN was evaluated on the PASCAL-VOC 2012 segmentation dataset consisting of 21 classes (20 foreground and 1 background class). The PASCAL 2012 segmentation dataset consists of 1464 (train) and 1449 (val) images for training and validation respectively.

~\autoref{table} shows the comparison of various CNN models trained on datasets listed in the first column. The performances reported are calculated according to the standard metric, mean of pixel-wise intersection-over-union (IoU). The first row lists the 21 classes and the mean IOU over all 21 classes.

The first row displays the performance when FCN is fine-tuned for real images with strong pixel-wise annotations from PASCAL VOC 2012 segmentation training dataset addressed as \texttt{Real(1.5k)}. Using bound-box annotated images as weak annotations was an alternative proposed earlier by ~\cite{Papandreou2015} which performed better than just using standard training dataset (with size of approximately 1.5k images). The second row showcases that the model trained with 10k weak bound-box annotated data (converted to pixel-wise labels using CRF) improved the mean-IoU performance from 47.68\% to 52.80\%. The predictions from CNN trained on \texttt{Weak(10k)} are represented in~\autoref{fig:Comp} third column. On comparing them with ground truth, it is observed that predictions miss the shape and sharp boundaries of the object.

Considering the CNN fine-tuned for \texttt{Weak(10k)} dataset as the baseline, we further fine-tuned it with rendered images of single class. ~\autoref{table} highlights the effect of using synthetic dataset on few classes namely car, bottle and aeroplane. \texttt{Syn\_Car(100)} denotes the dataset of 100 synthetic images with car as the object of interest. We observed that by using few synthetic images from single class, the performance of segmenting car as well as 7 other classes improved. The improvement in other classes can be explained by common features learned from the car images. This trend can be observed for other classes like bottle (\texttt{Syn\_Bot(100)}) and aeroplane (\texttt{Syn\_Aero(100)}).

Finally, we fine-tuned the baseline model with complete set of synthetic images (100 images per class; 20 classes) addressed as \texttt{Syn(2k)}. The mean IoU of this model increased from 52.80\% to 55.47\% as shown in ~\autoref{table} which clearly proves our hypothesis of supplementing synthetic images with weak annotated dataset. Some classes (car, bottle) showcased a significant improvement (10\% for car, 8\% for bottle) indicating the synthetic images in such cases to be more informative than others. While the classes like bicycle, dog, person and TV-monitor had lower IoU values since we had fewer 3D models available for those object types. Since objects like cow, cat, person etc. have highly variable appearance compared to other object classes, we observe lesser improvement in performance. 

To further explore the usefulness of synthetic and weak annotated dataset in conjunction with strong annotated real dataset, we fine-tune FCN with\\~\texttt{Real(1.5k)+Weak(10k)+Syn(2k)}. The model achieves 58.27\% (mean IoU) while\\~\texttt{Real(1.5k)+Syn(2k)} achieves 5.08\% (mean IoU) indicating the negative effect of non-photorealistic rendered images on strong annotation in real dataset.

~\autoref{fig:Comp} shows the comparison of the semantic labels generated from network trained on \texttt{Weak(10k)} dataset and \texttt{Weak(10k)+Syn(2k)} dataset. The latter predictions are better since they produce sharper edges and shapes. The results prove that shape information from the synthetic models help eliminate the noise generated from CRF in labels.
It is worth noting that even though synthetic images are non photo-realistic, and lack visual information from relevant backgrounds for objects, multiple object class in a single image or rich textures but they represent higher hierarchical feature like shape and thus can be used alongside weakly annotated images to achieve better performance on semantic segmentation tasks. The benchmark performance of FCN-8s on PASCAL test data when trained on augmented real image dataset released by ~\cite{Hariharan2011} with strong annotations is 62.2\% (mean IoU). While comparing with the benchmark performance, our model performs reasonably well with 55.47\% (mean IoU) trained with the total of 12k images  (\texttt{Weak(10k)+Syn(2k)}).

\section{Conclusion}\label{sec:conclusion}
Our report demonstrates a promising approach to minimize the annotation and dataset collection efforts by using rendered images from freely available 3D models. The comparison shows that using 10k weakly annotated images (which approximately equals the annotation efforts for 1.5k strong labels) with just 2k synthetic rendered images gives a significant rise in segmentation performance.

This work can be extended by training CNN with larger synthetic dataset, with richer 3D models and relevant backgrounds. Adding other features like relative scaling and occlusions can further strengthen the synthetic dataset. The effect of using synthetic dataset with improved architectures for semantic segmentation are being explored further. Further investigation can be done on factors like domain adaptation, co-adaptation among deeper layers that affect the transfer learning from synthetic to real images. 

\section*{Acknowledgment}
We acknowledge funding support from Innit Inc. consultancy grant\\ CNS/INNIT/EE/P0210/1617/0007.

\bibliographystyle{splncs}
\bibliography{synseg}

\end{document}